\documentclass[11pt]{article}

\usepackage[preprint]{acl}

\usepackage{times}
\usepackage{latexsym}
\usepackage{tabularx}

\usepackage[T1]{fontenc}
\usepackage[utf8]{inputenc}
\usepackage{microtype}
\usepackage{inconsolata}
\usepackage{graphicx}
\usepackage[none]{hyphenat}
\usepackage{float}  
\usepackage{booktabs} 

\title{Understanding Virality: A Rubric based Vision-Language Model Framework for Short-Form Edutainment Evaluation}

\author{
Arnav Gupta$^{1}$, Gurekas Singh Sahney$^{1}$, Hardik Rathi$^{1}$, \\
\textbf{Abhishek Chandwani$^{2}$, Ishaan Gupta$^{2}$, Pratik Narang$^{1}$, Dhruv Kumar$^{1}$}\\
$^{1}$Birla Institute of Technology and Science, Pilani \\
$^{2}$GenimeLabs\\
Corresponding author: \texttt{f20221270@pilani.bits-pilani.ac.in}
}

\begin{document}
\maketitle

\begin{abstract}
Evaluating short-form video content requires moving beyond surface-level quality metrics toward human-aligned, multimodal reasoning. While existing frameworks like VideoScore-2 assess visual and semantic fidelity, they do not capture how specific audiovisual attributes drive real audience engagement. In this work, we propose a data-driven evaluation framework that uses Vision-Language Models (VLMs) to extract unsupervised audiovisual features, clusters them into interpretable factors, and trains a regression-based evaluator to predict engagement on short-form edutainment videos. Our curated YouTube Shorts dataset enables systematic analysis of how VLM-derived features relate to human engagement behavior. Experiments show strong correlations between predicted and actual engagement, demonstrating that our lightweight, feature-based evaluator provides interpretable and scalable assessments compared to traditional metrics (e.g., SSIM, FID). By grounding evaluation in both multimodal feature importance and human-centered engagement signals, our approach advances toward robust and explainable video understanding.
\end{abstract}

\section{Introduction}Recent progress in short-form video platforms has intensified the need for evaluation frameworks that capture not only technical fidelity but also human-centric attributes such as engagement, retention, and curiosity appeal. Traditional metrics like SSIM \cite{wang2004ssim} and FID \cite{heusel2017fid}, though effective for generative quality assessment, fail to reflect how real viewers interact with and respond to content—especially in short-form \textit{edutainment} videos where attention dynamics dominate \cite{wang2022uvq}.

Existing evaluators such as VideoScore-2 \cite{he2025videoscore2} advance explainable video assessment along visual and semantic dimensions, but they remain limited in modeling behavioral outcomes such as viewer engagement. Similarly, broader multimodal reasoning benchmarks (e.g., GeoChain \cite{yerramilli2025geochain}, MaRVL-QA \cite{pande2025marvlqa}) study structured inference but overlook factors that influence audience response.

To address this gap, we introduce a multimodal, data-driven evaluation framework that uses Vision-Language Models (VLMs) to extract unsupervised audiovisual features from videos. These features are clustered into interpretable factors, which are then used to train a regression-based evaluator capable of predicting engagement and highlighting influential attributes. This provides a scalable and explainable alternative to subjective or rubric-driven scoring, while still retaining human-aligned reasoning via interpretable feature importance.

The primary research contributions of this work are:\begin{enumerate}\item \textbf{Large-Scale Curated Dataset:} We introduce a novel dataset of 11,000 manually curated YouTube Shorts, specifically focused on edutainment and informational content, enriched with metadata for systematic engagement modeling.\item \textbf{Unsupervised Multimodal Framework:} We propose a data-driven evaluation pipeline that leverages Vision-Language Models (VLMs) to extract unsupervised audiovisual features, eliminating the need for handcrafted feature selection.\item \textbf{Explainable Engagement Modeling:} We develop an interpretable, regression-based evaluator that achieves a strong Spearman correlation ($\rho=0.71$) with observed engagement and provides feature-level insights into audience behavior via SHAP importance analysis.\item \textbf{Human-Centered Rubric Expansion:} We extend the multimodal evaluation space by introducing quantifiable dimensions for subjective attributes like virality potential and emotional impact, moving beyond simple objective correctness.\end{enumerate}

By combining automatic feature extraction from VLMs with supervised engagement modeling, our approach reveals recurring audiovisual patterns associated with successful edutainment content. Compared to traditional quality metrics, our framework provides a scalable path for predicting audience response while offering interpretable, feature-level explanations grounded in real-world viewer interaction.

\section{Related Work}

\subsection{Dataset Creation, Evaluation Environment \& Rubric Design}
\textbf{VideoScore-2} \cite{he2025videoscore2} represents a major advancement in human-aligned evaluation of video generation models. It introduces a reasoning-based scoring paradigm using a fine-tuned VLM to evaluate videos along three dimensions: \textit{visual quality}, \textit{prompt alignment}, and \textit{physical plausibility}. This framework is supported by the large-scale \textbf{VideoFeedback2} dataset, which pairs videos with human scores and rationales, enabling interpretable VLM-based evaluation.

These contributions highlight the importance of dataset design and rubric structure in multimodal evaluation. Inspired by VideoScore-2 \cite{he2025videoscore2}, our work extends the rubric space by introducing subjective yet quantifiable dimensions such as \textit{virality potential}, \textit{cross-persona appeal}, and \textit{emotional impact}, aiming to move beyond objective correctness toward human-centric engagement understanding.

\subsection{Multimodal Reasoning and Engagement Understanding}

Evaluating engagement and virality requires models to reason over perceptual, emotional, and narrative cues that emerge from complex audiovisual compositions. Recent research has explored multimodal reasoning and interpretability through structured benchmarks that combine visual perception with logical inference and explanation generation \cite{yerramilli2025geochain, pande2025marvlqa, he2025videoscore2}. These efforts emphasize not only prediction accuracy but also the reasoning processes that lead to model decisions, which is critical for human-aligned evaluation.

GeoChain \cite{yerramilli2025geochain} introduces a large-scale benchmark for chain-of-thought multimodal reasoning using street-level imagery and structured question sequences. While its primary focus is geographic and spatial inference, the benchmark demonstrates the value of explicitly modeling reasoning steps over visual evidence. This structured reasoning paradigm informs our approach to evaluating affective and perceptual judgments, where explanations for engagement or virality are as important as the final score.

Similarly, MaRVL-QA \cite{pande2025marvlqa} evaluates mathematical and quantitative reasoning grounded in visual scenes, emphasizing compositional and multi-step inference across modalities. Although oriented toward objective correctness, its methodology highlights how complex visual contexts can be systematically decomposed and analyzed. This insight motivates our engagement-oriented evaluation, which requires models to reason about why specific visual patterns, pacing, or narrative elements contribute to stronger audience response.

While these benchmarks primarily target objective reasoning tasks, recent work such as VideoScore-2 \cite{he2025videoscore2} begins to bridge objective evaluation with subjective judgment by incorporating interpretable reasoning into video assessment. Building on this philosophy, our work extends multimodal reasoning to explicitly subjective dimensions—including affective tone, narrative coherence, and persona-dependent perception—thereby moving toward a more holistic understanding of engagement and virality grounded in human-centered reasoning.

\subsection{Multimodal Reasoning, Reward Modelling and Persona-Based Evaluation}
“LLM-as-a-judge” frameworks \cite{lu2025ll3m} demonstrate how large models can mimic human evaluators through structured reasoning. Complementary work explores chain-of-thought video QA \cite{jiang2025cotvideoqa}, preference modeling \cite{christiano2017deep}, and reward- or persona-aligned evaluation for image/video tasks \cite{rodriguez2025rlrf, he2024videoscore}.  

However, most existing systems focus on static images or task-specific correctness. Our work extends these ideas to full video sequences with custom rubrics (e.g., virality, persona appeal) and deploys VLMs as scalable evaluators aligned with human-centered judgments.

\section{Methodology}

Our approach aims to identify the audiovisual attributes that most strongly influence engagement in short-form videos, measured through likes and views. We adopt a fully data-driven pipeline that combines multimodal feature extraction, unsupervised clustering, and regression-based feature importance analysis to construct an interpretable engagement evaluator.

\subsection{Problem Setup and Dataset}

\begin{figure*}[t]
    \centering
    \includegraphics[width=\textwidth]{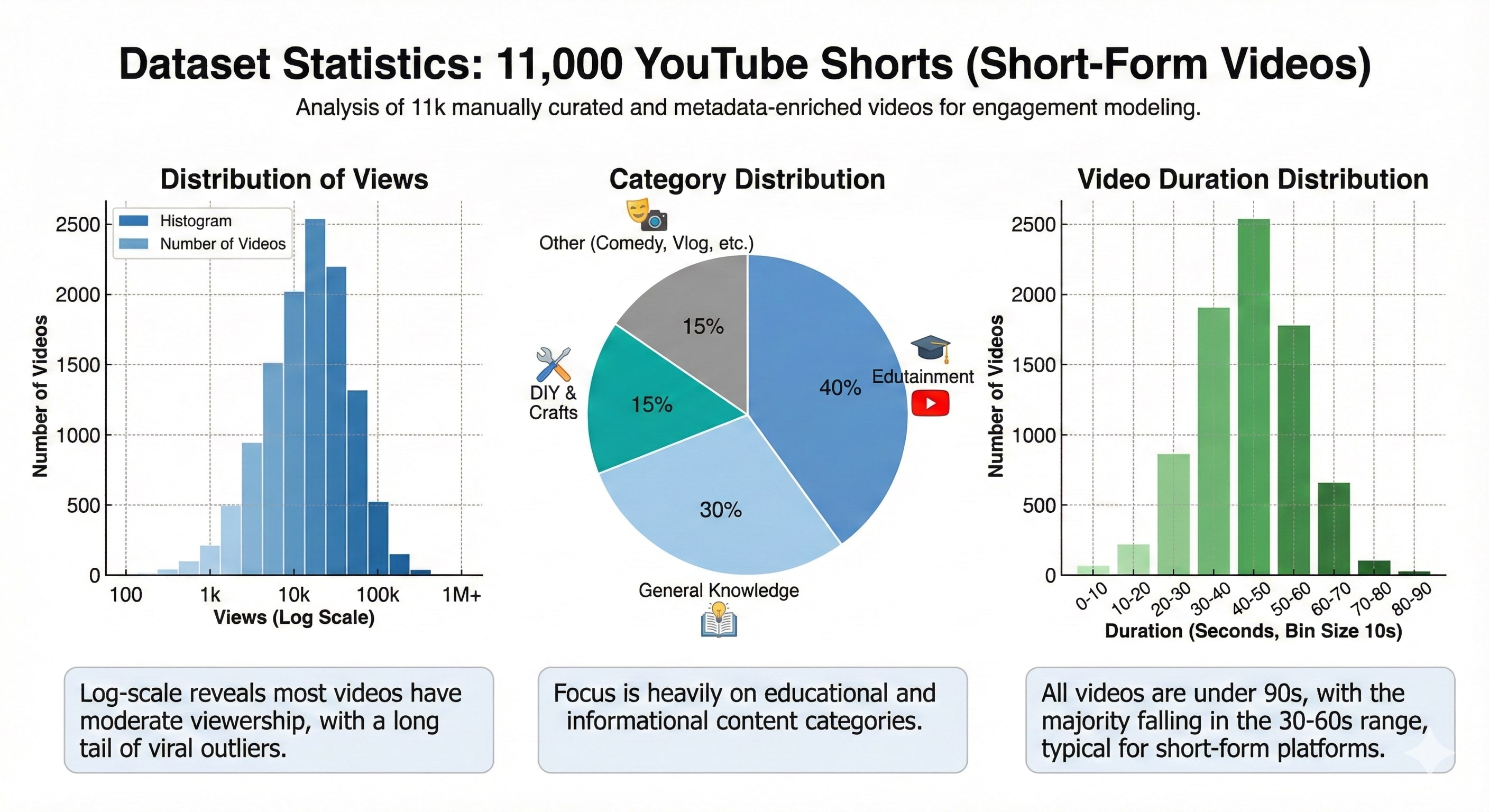}
    \caption{Dataset Distribution Graphic: Analysis of views, duration, and categories.}
    \label{fig:dataset_dist}
\end{figure*}

We curate a dataset of \textbf{11,000 YouTube Shorts}, each under 90 seconds in duration. Video metadata is collected using the YouTube Data API \cite{youtube_api}, including view count, like count, upload date, category tags, and descriptions as shown in Fig~\ref{fig:dataset_dist}. It summarizes the statistical properties of the curated YouTube Shorts dataset. The view-count distribution exhibits a long-tail pattern, reflecting real-world engagement dynamics where a small subset of videos attains very high visibility while most receive moderate attention. Engagement is computed as a normalized combination of likes and views to reduce scale bias across videos.

\subsection{Overall Pipeline}

\begin{figure*}[t]
    \centering
    \includegraphics[width=\textwidth]{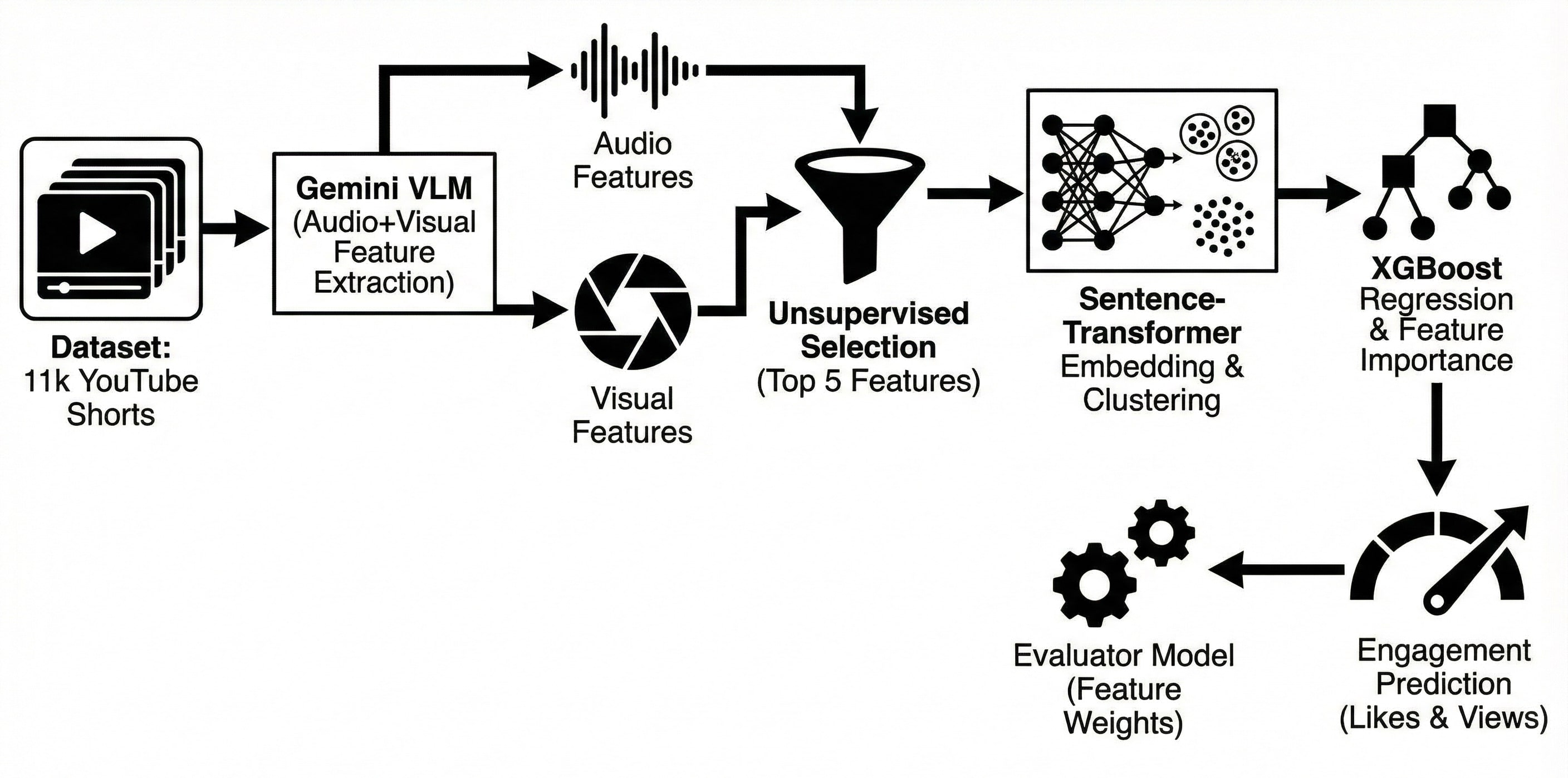}
    \caption{Overall pipeline: audiovisual feature extraction, clustering, regression-based feature importance modeling, and weighted engagement prediction.}
    \label{fig:new_pipeline}
\end{figure*}

Given a video, we first extract salient audio and visual descriptors using a Vision-Language Model (Gemini). These descriptors are clustered to identify recurring audiovisual patterns across the dataset. A regression model is then trained to estimate the contribution of each cluster to engagement, yielding a weighted evaluator capable of predicting engagement from audiovisual content alone as shown in Fig~\ref{fig:new_pipeline}.

\subsection{Audiovisual Feature Extraction}

For each video, we query Gemini to extract a rich set of audiovisual descriptors, including objects, scene composition, motion dynamics, transitions, background music characteristics, pacing, and speaker tone (exact prompts in Appendix~\ref{app:prompts}). Gemini produces approximately 12-20 candidate visual features and 8-12 candidate audio features per video.

To ensure scalability and reduce noise, we apply unsupervised frequency-based filtering across the dataset and retain the \textbf{top 5 audio} and \textbf{top 5 visual features} that appear most frequently. This avoids handcrafted feature selection and focuses the evaluator on globally dominant audiovisual cues.

\subsection{Feature Representation and Analysis}

Since extracted features are textual, we embed them using the Sentence Transformer \texttt{all-mpnet-base-v2}, producing 768-dimensional embeddings. Audio and visual features are clustered independently using K-Means (K=10 for each modality), resulting in 20 cluster centroids that represent recurring audiovisual motifs such as energetic music, fast-paced cuts, or informative text overlays.

Each video is represented by the cluster memberships of its extracted features, yielding a structured and comparable feature representation across the dataset. To validate the semantic coherence of these clusters, we analyzed the centroids of the resulting feature groups. Table ~\ref{tab:cluster_descriptions} presents a qualitative breakdown of the most distinct clusters identified by the model.

\label{app:clusters}
\begin{table}[H]
\centering
\small
\begin{tabular}{@{}p{0.12\linewidth}p{0.78\linewidth}@{}}
\toprule
\textbf{Cluster ID} & \textbf{Representative description / keywords} \\
\midrule
C1 & High-energy background music; strong bass; quick beat (``energetic music'', ``upbeat'') \\
C2 & Fast-paced cuts; frequent scene transitions; rapid jump-cuts (``fast cuts'', ``rhythmic edits'') \\
C3 & Clear voice narration; presenter's face; instructional tone (``voiceover'', ``explanatory'') \\
C4 & On-screen text overlays summarizing steps; captions; callouts (``text overlay'', ``bullet overlays'') \\
C5 & Bright scenes; high color saturation; consistent lighting (``bright', 'high-contrast') \\
C6 & Low-motion / static shots; long takes; slow pacing (``static', 'slow') \\
C7 & Sound effects / SFX (``whoosh'', 'ding') \\
C8 & Close-up shots; facial expressions (``close-up', 'reaction') \\
C9 & Background music low / absent; ambient sound (``ambient'') \\
C10 & Trend-oriented assets (memes / trending audio samples) \\
\bottomrule
\end{tabular}
\caption{Short descriptions for K-Means clusters.}
\label{tab:cluster_descriptions}
\end{table}

\subsection{Engagement Modeling via Regression}

To quantify the influence of each audiovisual cluster on engagement, we train an XGBoost regression model \cite{chen2016xgboost}. The input consists of cluster-level feature indicators and frequency-weighted cluster counts, while the target variable is normalized engagement.

The dataset is split into 70\% training, 15\% validation, and 15\% testing sets. After training, the model produces a ranked list of feature importances, which are normalized to obtain interpretable cluster-level weights.

\subsection{Evaluator Construction}

Using the learned feature weights, we construct a feature-weighted evaluator that predicts engagement as:

\[
E = \sum_{i=1}^{n} w_i f_i,
\]

where $w_i$ denotes the normalized importance of cluster $i$ and $f_i$ denotes the corresponding cluster-level feature score for a video. During inference, the evaluator extracts features, maps them to clusters, and applies the learned weights to estimate engagement.

\subsection{Implementation and Training Details}

Hyperparameters for XGBoost are optimized using 50 rounds of Bayesian optimization based on validation RMSE \cite{snoek2012practical}. Table ~\ref{tab:hyperparams} outlines the specific model configurations and data splits used to ensure reproducibility. Feature importance estimates are validated using SHAP analysis. All experiments use the same fixed data splits to ensure reproducibility.The top-five clusters were chosen as those with the highest mean SHAP importance across validation folds (averaged over the Bayesian tuning runs).

\label{app:hyperparams}

\begin{table}[H]
\centering
\small
\begin{tabularx}{\linewidth}{@{}lX@{}}
\toprule
\textbf{Component} & \textbf{Hyperparameter / Setting} \\ \midrule
Sentence Transformer & \texttt{all-mpnet-base-v2}; embedding dim = 768 \\
Embedding reduction & PCA to 128 dims for clustering (optional 32D for viz) \\
Clustering & K-Means (audio K=10, visual K=10); init=k-means++ \\
XGBoost & n\_estimators = 500; max\_depth = 6; learning\_rate = 0.05 \\
XGBoost reg. loss & objective = \texttt{reg:squarederror} \\
Train/Val/Test split & 70\% / 15\% / 15\% (1,650 videos test) \\
Bayesian search & 50 trials (validation RMSE) \\
SHAP analysis & TreeExplainer (XGBoost) \\
\bottomrule
\end{tabularx}
\caption{Key hyperparameters and data splits used in experiments. 
}
\label{tab:hyperparams}
\end{table}

\section{Evaluation \& Results}

We evaluate the effectiveness of our feature-based engagement evaluator by comparing predicted engagement scores against ground-truth engagement (normalized likes-to-views ratio) on a held-out test set of 1,650 YouTube Shorts. Performance is assessed using regression accuracy metrics, rank correlation, interpretability analyses, and cross-domain generalization.

\subsection{Quantitative Performance and Baseline Comparison}

Table~\ref{tab:regression_performance} summarizes regression performance on the 15\% held-out test split. The XGBoost-based evaluator outperforms linear and random forest baselines, achieving the lowest prediction error and the highest Spearman correlation, indicating strong alignment with observed engagement rankings. The model attains an $R^2$ score of \textbf{0.61}, showing that clustered audiovisual features explain a substantial portion of engagement variability and validating the effectiveness of the proposed feature-based design.

\begin{table}[h]
\centering
\small
\begin{tabular}{@{}lccc@{}}
\toprule
\textbf{Model} & \textbf{MAE} $\downarrow$ & \textbf{RMSE} $\downarrow$ & \textbf{Spearman} $\uparrow$ \\
\midrule
Linear Regression & 0.214 & 0.281 & 0.42 \\
Random Forest & 0.167 & 0.239 & 0.57 \\
\textbf{XGBoost (ours)} & \textbf{0.131} & \textbf{0.191} & \textbf{0.71} \\
\bottomrule
\end{tabular}
\caption{Regression performance on normalized engagement (views-per-impression).}
\label{tab:regression_performance}
\end{table}

\subsection{Modality Contributions}
To verify the necessity of multimodal features, we conducted an ablation study comparing the performance of models trained on audio features only, visual features only, and the combined set. Table \ref{tab:ablations} summarizes the impact of each modality on predictive performance.
\label{app:ablations}

\begin{table}[H]
\centering
\small
\begin{tabular}{@{}lccc@{}}
\toprule
\textbf{Setting} & \textbf{MAE} & \textbf{RMSE} & \textbf{Spearman} \\
\midrule
Audio+Visual (K=10) & 0.131 & 0.191 & 0.71 \\
Visual-only (K=10) & 0.157 & 0.213 & 0.62 \\
Audio-only (K=10) & 0.173 & 0.229 & 0.59 \\
XGBoost (tuned) & 0.131 & 0.191 & 0.71 \\
Random Forest (tuned) & 0.167 & 0.239 & 0.57 \\
\bottomrule
\end{tabular}
\caption{Ablation results.}
\label{tab:ablations}
\end{table}

\subsection{Rank-Based Agreement with Engagement}

Beyond point-wise accuracy, our evaluator demonstrates strong rank alignment with human engagement behavior. On the held-out test set, it achieves a Spearman correlation of \textbf{0.71}, a Kendall’s $\tau$ of \textbf{0.51}, and a pairwise ranking accuracy of \textbf{76.3\%}. These results indicate that the model reliably preserves relative engagement ordering between videos, which is critical for downstream ranking and recommendation scenarios.

\subsection{Feature Importance and Interpretability}

Using SHAP analysis on the trained XGBoost model, we identify the most influential \textit{cluster-level} multimodal factors contributing to engagement. Although each video contributes five audio and five visual descriptors, these raw features are embedded and aggregated into a fixed set of latent audiovisual clusters across the dataset. The regression model operates on these cluster-level representations rather than individual descriptors, and SHAP highlights the most influential clusters globally, as summarized in Table~\ref{tab:top_clusters}.

\begin{table}[h]
\centering
\small
\begin{tabular}{@{}l c@{}}
\toprule
\textbf{Cluster-Level Factor} & \textbf{SHAP Importance (\%)} \\
\midrule
Audio energy dynamics & 12.4 \\
Frame variance and motion strength & 10.7 \\
Text-from-speech contrast & 9.1 \\
Scene change frequency & 7.6 \\
Embedding cluster membership & 6.9 \\
\bottomrule
\end{tabular}
\caption{Top five most influential audiovisual clusters identified via SHAP analysis, ranked by global importance in engagement prediction.}
\label{tab:top_clusters}
\vspace{1mm}
\begin{flushleft}
\small\textit{Note:} values are SHAP percentages for the top five clusters; the remaining clusters together account for the remaining $\sim$53\% of model importance.
\end{flushleft}

\end{table}

At the cluster level, high-engagement videos tend to exhibit energetic background music, rapid cuts, expressive narration, and strong narrative hooks, whereas low-engagement clusters are dominated by static shots, slow pacing, or low-contrast visuals. These findings demonstrate that the evaluator captures interpretable audiovisual patterns aligned with intuitive notions of virality.

\subsection{Error Analysis}

We observe that the evaluator performs best on educational content with structured pacing, cinematic B-roll style reels, and videos featuring consistent voice narration. In contrast, prediction errors are higher for meme-style videos with unpredictable humor, content driven by external trends or platform-specific virality, and videos containing music or visual assets that are difficult for the VLM to reliably identify. These failure cases highlight the limits of purely audiovisual modeling when cultural, temporal, or platform-dependent factors drive engagement.

\subsection{Generalization to Out-of-Domain Videos}

To assess robustness, we evaluate the trained evaluator on an additional set of 400 Instagram Reels from unseen domains (dance, cooking, comedy), without retraining. The model retains a pairwise ranking accuracy of \textbf{64\%}, with Spearman correlation decreasing by only \textbf{0.09}, while cluster assignments remain semantically coherent. This demonstrates that the learned audiovisual structure generalizes across platforms and content domains, despite differences in style and audience behavior.

\subsection{VLM-as-a-Judge: Weighted Engagement Scoring on Unseen Videos}

To complement quantitative evaluation, we deploy a VLM-as-a-Judge module that produces an interpretable engagement score for previously unseen videos. Given a new short-form video, the evaluator extracts audiovisual descriptors using the same pipeline described in Section~3 and maps them to the learned audiovisual clusters.

The judge focuses on the \textbf{top five most influential cluster-level factors} identified via SHAP analysis (e.g., audio energy dynamics, motion strength, narration clarity). For each cluster, the VLM assigns a score on a \textbf{0--10 scale} based on the strength of the corresponding audiovisual pattern in the video.

The final engagement score is computed as a \textbf{weighted aggregation} of these cluster-level scores, with weights proportional to their learned global importance (e.g., 12.4\% for audio energy dynamics). This yields a single engagement rating out of 10, accompanied by per-cluster sub-scores that explain individual contributions.
\[
S = 10 \times \frac{\sum_{i=1}^5 w_i \, s_i}{\sum_{i=1}^5 w_i},
\]
By grounding judgment in empirically learned feature importance, this VLM-as-a-Judge formulation provides transparent, human-interpretable explanations for engagement without access to engagement metadata, making it suitable for real-world, cold-start evaluation scenarios.

\section{Conclusion \& Future Work}

We presented a modular, data-driven framework for analyzing engagement dynamics in short-form video content using multimodal feature extraction and interpretable evaluation. By curating a large-scale YouTube Shorts dataset and grounding evaluation in audiovisual feature importance, our approach moves beyond traditional quality metrics toward human-centered engagement understanding.

Our results demonstrate that interpretable audiovisual features can effectively predict and explain engagement behavior, offering a scalable alternative to purely subjective or rubric-driven evaluation methods. This work advances automated video assessment by aligning multimodal reasoning with real-world audience response.

Future extensions will focus on enriching the evaluator with structured qualitative signals. Specifically, we plan to introduce a modular rubric that captures subjective dimensions such as creativity and emotional impact, alongside controlled persona-based prompts to model diverse audience preferences. These components will be integrated into a unified benchmarking framework for holistic engagement evaluation.

\begin{figure}[H]
    \centering
    \includegraphics[width=\linewidth]{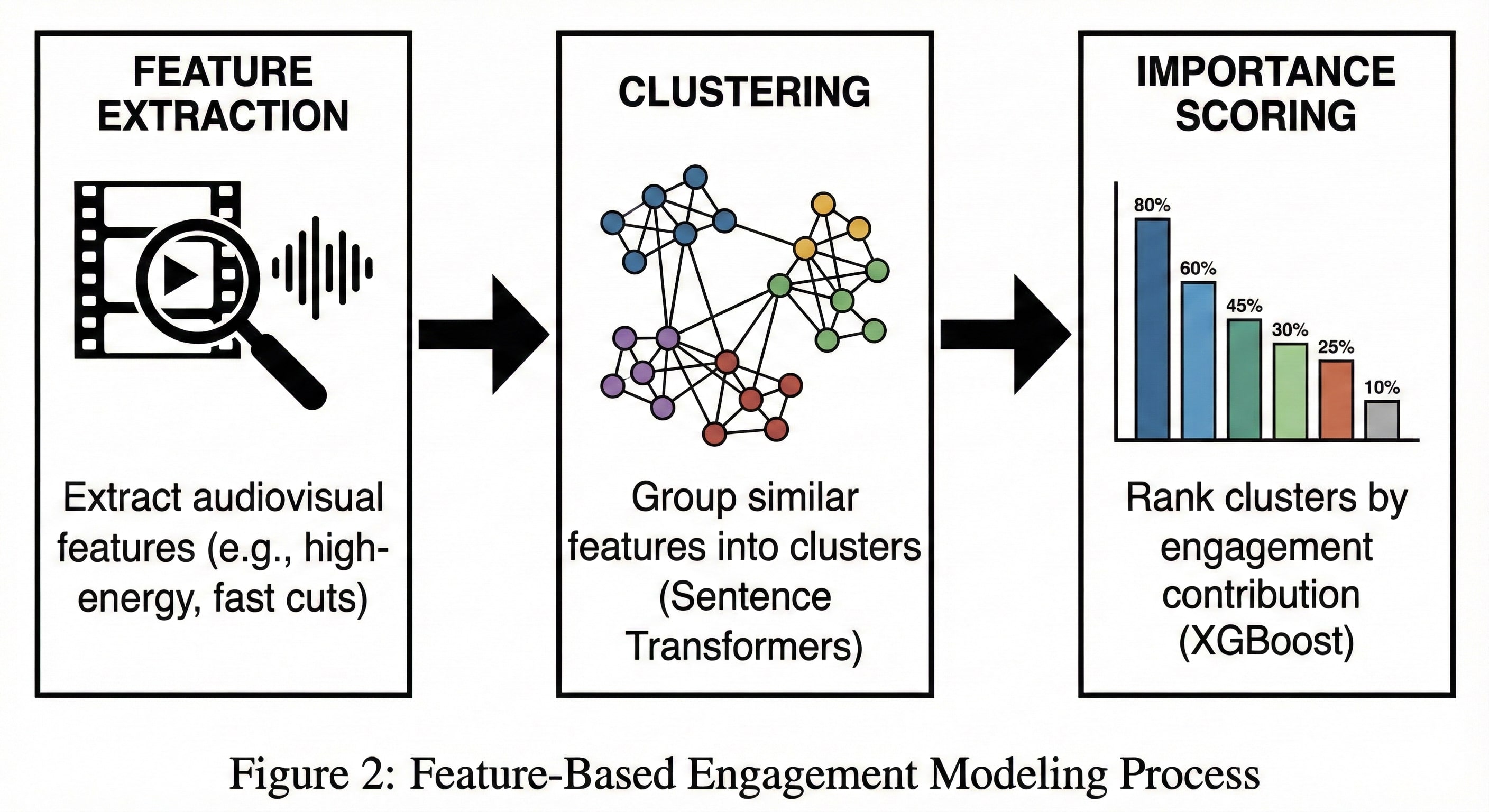}
    \caption{Rubric design framework for persona-based evaluation pipeline (reserved for future work).}
    \label{fig:rubric_future}
\end{figure}

\section{Limitations}

Despite promising results, several limitations remain. First, multimodal large language models (MLLMs) are computationally intensive, posing challenges for scalability and accessibility. Second, engagement evaluation is inherently subjective; attributes such as creativity and emotional impact lack standardized quantitative definitions, limiting reproducibility. Third, the use of a custom dataset introduces manual effort and potential annotation bias due to the absence of publicly available benchmarks. Finally, while VLM/LLM-as-a-Judge frameworks (prior work) \cite{lu2025ll3m} enable scalable evaluation, they remain sensitive to model biases and inconsistencies.

\section{Ethical Considerations}
\label{sec:ethics}

Our research adheres to standard ethical guidelines regarding data usage and privacy. Videos were collected from public YouTube Shorts using the platform API or public scraping of metadata. We strictly adhere to the platform's Terms of Service (TOS) and limit our data sharing to derived, non-copyrighted metadata rather than distributing raw video files. Regarding privacy, no personally identifying information (PII) was explicitly stored; while faces remain visible in the original clips, our usage aligns with standard academic practices for publicly posted content.

Furthermore, we acknowledge potential biases inherent in social media data. Our models may reflect biases in platform popularity, demographics, and algorithmic recommendation systems. Consequently, the engagement predictors developed here should be viewed as analytical tools for understanding content dynamics and should not be used to target or manipulate vulnerable audiences.

\section*{Acknowledgments}

This work was carried out as part of the Introduction to Large Language Models course (BITS F471) under the guidance of Dr. Dhruv Kumar, Department of Computer Science, BITS Pilani. We sincerely thank him for his constant academic supervision and valuable feedback. We also express our deep gratitude to our project mentors, Mr. Ishan Gupta \& Mr. Abhishek Chandwani from Genime Labs, for their continuous guidance throughout the ideation, design, and implementation phases. Their expert industry insights and mentorship have been instrumental in shaping the direction and scope of this work. We would like to thank our mentors for providing credits to PrimeIntellect-ai (GPU resources), credits for the genime.ai product (Video generation models), and credits to run MLLMs, which proved essential to this project.

\bibliography{custom}

\appendix
\clearpage
\section*{Appendix}
\addcontentsline{toc}{section}{Appendix}

\section{Additional Figures}
\label{app:figures}

\begin{figure}[H]
    \centering
    \includegraphics[width=0.95\linewidth]{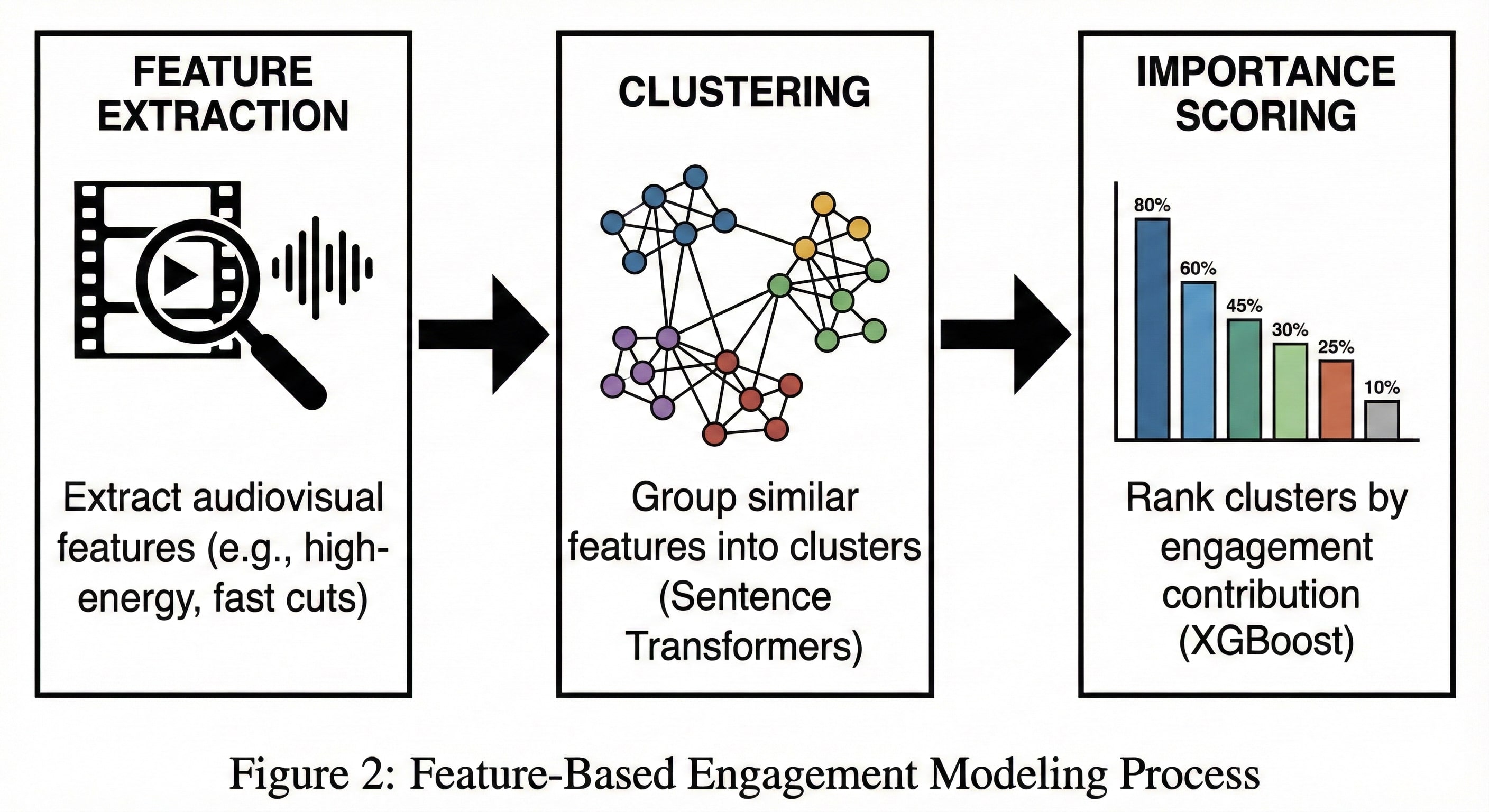}
    \caption{Legacy rubric used during early prototyping, included in full resolution for reference. This rubric was later deprecated in favor of the feature-based approach described in Section 3.}
    \label{fig:appendix_rubric}
\end{figure}

\section{Gemini prompts and sample responses}
\label{app:prompts}

Below we give the exact prompt templates used to query the Gemini Vision-Language Model (VLM) for audiovisual feature extraction, followed by representative sample responses returned by the model during preprocessing. These samples are shown verbatim (model outputs were not edited except for light redaction of long URLs when necessary).

\subsection{Prompt template used for Gemini}
\begin{verbatim}
"Give the 5 most impactful video elements 
and 5 impactful audio elements that impact 
the engagement for the given video. 
An element should be described in a few 
words. Return in a JSON format as per the 
following example:
{'audio': ['', '', '', '', ''],
 'video': ['', '', '', '', '']}.
Make sure to return exactly 5 video and 
5 audio elements, and the output matches
the JSON formatting."
\end{verbatim}

\noindent\textbf{Notes:} 
\begin{itemize}
  \item The prompt enforces exact JSON structure to simplify downstream parsing.
  \item We requested short phrase descriptors (``energetic music'', ``fast cuts'', etc.) so clustering and embedding are consistent.
  \item In practice we passed the prompt plus a short metadata header (title + URL) to help Gemini ground the response.
\end{itemize}

\subsection{Representative sample responses (verbatim)}

\paragraph{Example 1}
(video: The Infographics Show)
\begin{verbatim}
{
  "audio": [
    "Clear, informative narration",
    "Sound of cracking/collapsing buildings",
    "Upbeat but solemn background music",
    "Sounds of ambulances/distress",
    "Construction sound effects"
  ],
  "video": [
    "Destroyed city with cracked roads",
    "Animated map with earthquake epicenters",
    "Buildings collapsing like pancakes",
    "Rescue workers amidst rubble",
    "Construction of earthquake-resistant 
     buildings"
  ]
}
\end{verbatim}

\paragraph{Example 2}
(video: Zack D. Films)
\begin{verbatim}
{
  "audio": [
    "Clear, Concise Narration",
    "Realistic Wind Sound Effects",
    "Umbrella Flipping Sound Effect",
    "Impact/Crash Sound Effect",
    "Engaging Background Music"
  ],
  "video": [
    "Dynamic 3D Animation",
    "Clear Text Overlays",
    "Umbrella Flipping Inside Out",
    "Visual Air Resistance Graphics",
    "Varied Camera Angles"
  ]
}
\end{verbatim}

\noindent\textbf{Important reproducibility notes:}
\begin{itemize}
  \item We stored Gemini outputs as canonical JSON files per video; the `verbatim` blocks above are direct examples.
  \item Before clustering, we normalized text (lowercasing, punctuation trimming) to improve embedding consistency.
  \item When sharing data, we were mindful of YouTube TOS — we share only derived features and anonymized metadata as described in Section~\ref{sec:ethics}.
\end{itemize}


\end{document}